\documentclass[10pt,twocolumn,letterpaper]{article}

\usepackage{3dv}
\usepackage{times}
\usepackage{epsfig}
\usepackage{graphicx}
\usepackage{amsmath}
\usepackage{booktabs}
\usepackage[font=small]{caption}
\usepackage[font=normalsize]{subcaption}
\usepackage{amssymb}
\usepackage[dvipsnames]{xcolor}
\usepackage{pdflscape}

\usepackage[pagebackref=true,breaklinks=true,colorlinks,bookmarks=false]{hyperref}

\definecolor{gtgreen}{rgb}{0, 0.867, 0.391}
\definecolor{cfblue}{rgb}{0, 0.703, 0.859}
\definecolor{hmorange}{rgb}{1, 0.25, 0.06}
\definecolor{hmorange}{rgb}{1, 0.25, 0.06}
\definecolor{darkorange}{rgb}{0.8, 0.4, 0}

\makeatletter
\def\thanks#1{\protected@xdef\@thanks{\@thanks
        \protect\footnotetext{#1}}}
\makeatother

\threedvfinalcopy

\def\threedvPaperID{310}

\ifthreedvfinal\fi
\begin{document}

\title{Reconstructing and grounding narrated instructional videos in 3D}

\author{Dimitri Zhukov$^{1*}$ \and Ignacio Rocco$^{1,2*}$ \and Ivan Laptev$^{1}$ \and Josef Sivic$^{3}$ \and Johannes L. Schönberger$^{4}$ \and Bugra Tekin$^{4}$ \and Marc Pollefeys$^{4,5}$ \thanks{$^*$Equal contribution} \thanks{$^1$Inria, Ecole normale suprieure, CNRS, PSL Research University, Paris, France} \thanks{$^2$Now at Facebook AI Research} \thanks{$^3$CIIRC -- Czech Institute of Informatics, Robotics and Cybernetics at the Czech Technical University in Prague} \thanks{$^4$Microsoft} \thanks{$^5$ETH Zurich}}

\maketitle

\begin{abstract}
Narrated instructional videos often show and describe manipulations of similar objects, e.g., repairing a particular model of a car or laptop. In this work we aim to reconstruct such objects and to localize associated narrations in 3D.
Contrary to the standard scenario of instance-level 3D reconstruction, where \emph{identical} objects or scenes are present in all views,
objects in different instructional videos may have large appearance variations 
given varying conditions and versions of the same product. Narrations may also have large variation in natural language expressions.
We address these challenges by three contributions. First, we propose an approach for correspondence estimation
combining learnt local features and dense flow.
Second, we design a two-step \emph{divide and conquer} reconstruction approach where the initial 3D reconstructions of individual videos are combined into a 3D alignment graph.
Finally, we propose an unsupervised approach to ground natural language in obtained 3D reconstructions. We demonstrate the effectiveness of our approach for the domain of car maintenance. Given raw instructional videos and no manual supervision, our method successfully reconstructs engines of different car models and associates textual descriptions with corresponding objects in 3D.
\end{abstract}

\section{Introduction}
Imagine you need to perform a task on a product you own, such as replacing the drum unit in your laser printer or re-filling the coolant fluid tank in your car, but you don't know how to proceed. In the past few years, instructional videos have become a popular internet resource for learning how to perform such kind of tasks. However, finding the answer to this particular problem would require searching for the appropriate video, identifying the key steps, and finally, putting it into practice by performing the right sequence of actions involving the correct object parts at the right locations. 
Wouldn't it be great to have a virtual assistant guide you through the steps in augmented reality showing you  \emph{which} tasks to perform and \emph{where} to perform them? In order to produce such an assistant, a method for relating natural language to the actual 3D objects in the world is needed. In this work we show that it is possible to obtain 3D reconstructions
and to learn associations between natural language and corresponding 3D locations (3D language grounding) using raw instructional videos and no manual annotations.

\begin{figure}[t]
\centering
\includegraphics[width=\columnwidth]{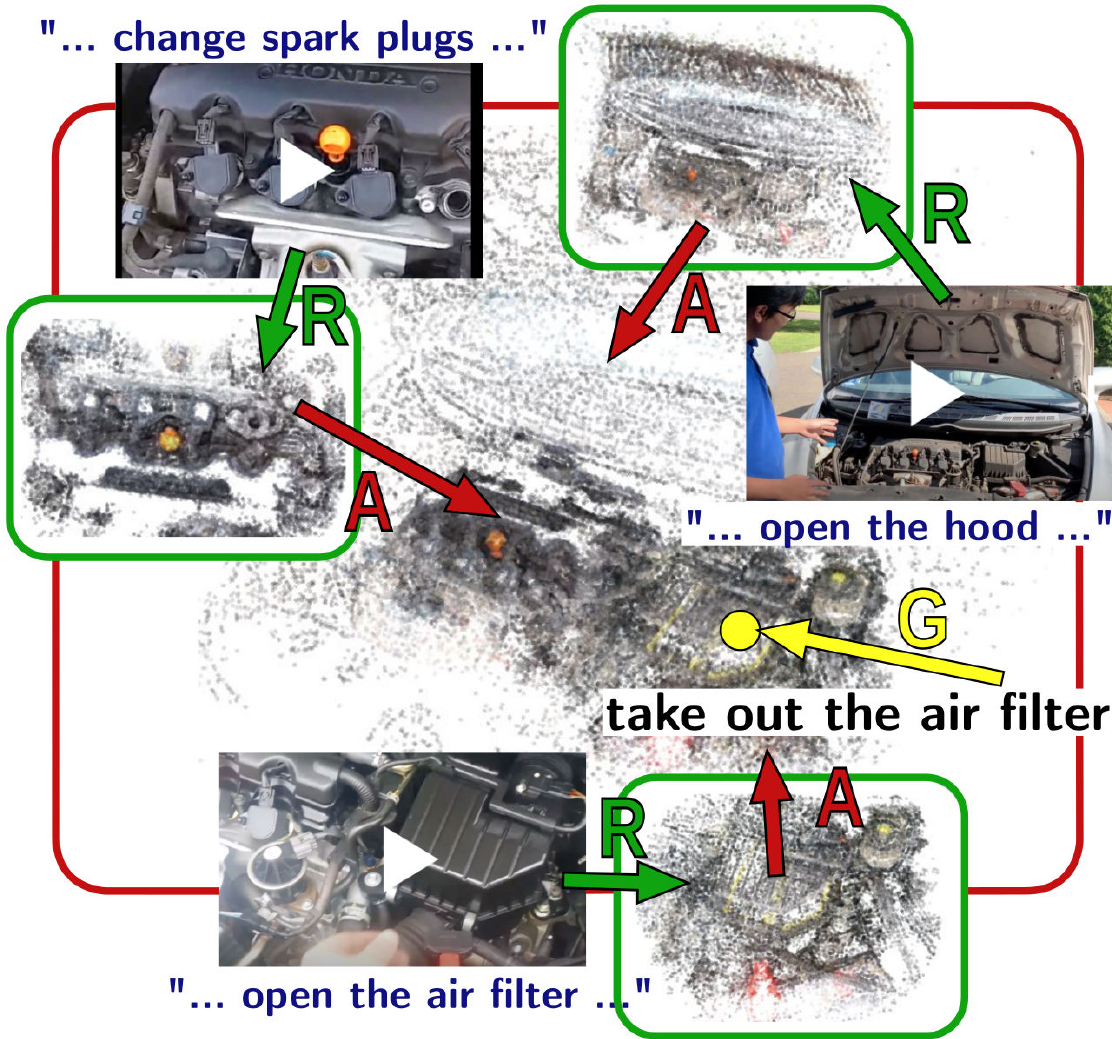}
\caption{\textbf{Illustration of the proposed method.} Given a set of instructional videos showing the same object (e.g. same car model), we propose a two-step 3D reconstruction approach where, first, each video is independently reconstructed (R) and, second, reconstructions are aligned (A) in a single model. Using the reconstructed 3D model together with the narrations from each individual video
and no manual supervision, we learn a language grounding model (G) that can determine the corresponding 3D region for a given text query such as ``air filter".
}
\vspace{-.3cm}
\label{fig:teaser}
\end{figure}

In the past, instructional videos have proven to be a useful source of data for 
learning to temporally localize a set of actions in a video strip, given the set of action labels (i.e.~weakly-supervised temporal action localization)~\cite{Malmaud15what,alayrac:hal-01580630,Huang2016connectionist,richard17weakly,richard18actionsets,sener18unsupervised,Kukleva_2019_CVPR,Zhukov_2019_CVPR,elhamifar20}. This is useful for downstream tasks such as video summarization or video retrieval.
However, with the advent of smartphones and headsets equipped with 3D sensors 
the problem of spatial action localization in 3D gains relevance~\cite{chen20153d, martin2017virtual, pratt2018through,kim2019foveated}. While raw video data is useful to learn visual representations of actions, it is difficult to directly learn about 3D structure from 2D frame data. Therefore, in order to solve the problem of spatial action localization in 3D, a suitable 3D spatial representation of the object and the scene are necessary. However, producing a 3D reconstruction from a set of narrated instructional videos presents several challenges.

In recent years, significant progress has been made in the field of 3D reconstruction. Structure-from-motion (SfM) pipelines~\cite{schonberger2016structure,moulon2016openmvg,wu2013towards,theia-manual,Moulon2012} leverage robust local features for obtaining correspondences between images despite large illumination and viewpoint changes. This has allowed to perform 3D reconstruction from unstructured photo-collections from the web, where images may be taken months or years apart~\cite{snavely2006photo, frahm2010building, agarwal2011building}. However, the task of reconstructing a set of narrated instructional videos poses some additional challenges. First, besides the large illumination and viewpoint changes, these videos may contain significant appearance changes. This is due to the fact that, while the objects that appear in different videos belong to the same brand and model, they might present variations due to different options or sub-versions of the same product. Therefore, obtaining correspondences between such videos is an intermediate problem between traditional instance-level matching (the \emph{same} physical object is imaged)~\cite{lowe1999object, hamming, snavely2006photo} and category-level matching (semantically related objects are imaged, such as two different species of dogs)~\cite{ham2016proposal,rocco2018end,rocco2018neighbourhood,min2019hyperpixel}. Despite the robustness of the local descriptors used in current SfM pipelines, these are not designed to handle such strong appearance changes and therefore are prone to failure. As a second challenge, because the computational cost of 3D reconstruction grows quadratically with the number of total frames, jointly reconstructing a set of many videos may be unfeasible, or may impose a strong limit in the number of frames to be sampled from each video.

In this work we address the above difficulties and propose a method that is (i) capable of finding correspondences across instructional videos despite the large appearance changes, and (ii) reduces the global computational complexity of 3D reconstruction by reconstructing each video independently and then combining the results. 
Given the obtained 3D reconstructions and corresponding video narrations, we next learn 3D language grounding.
Our approach to 3D language grounding leverages both narrations and visual information in instructional videos to associate the input text to different parts of the 3D scene, and doesn't require manual supervision (see Fig.~\ref{fig:teaser}).

\paragraph{Contributions.} In summary, our contributions are the following. \textbf{I.}\ We develop an approach for correspondence estimation that combines the strengths of local features and dense flow in order to obtain reliable matches across instructional videos. \textbf{II.}\ We propose a two-step 3D reconstruction approach that reduces the overall computational complexity by reconstructing each video independently, and then combining the results into a 3D alignment graph as a proxy to a single aligned 3D reconstruction. We evaluate obtained reconstructions on the task of keypoint transfer.  \textbf{III.}\,\, We propose an unsupervised method, that makes use of the obtained 3D models and narrations in instructional videos to learn language grounding in 3D.

\section{Related work}
\paragraph{3D reconstruction.}
The standard approach to obtain a 3D reconstruction from a set of images is to employ an incremental or global structure-from-motion pipeline such as COLMAP~\cite{schonberger2016structure}, Theia~\cite{theia-manual}, VisualSfM~\cite{wu2013towards} or OpenMVG~\cite{moulon2016openmvg}. In all cases, the first step is to find correspondences between the set of input images. This is typically done using local features such as SIFT~\cite{lowe1999object}, which are robust to moderate changes in illumination and viewpoint present in instance-level 3D reconstruction. In this work, we propose a different 3D reconstruction approach targeted to the reconstruction of narrated instructional videos, which have the additional challenge of large appearance changes due to differences in sub-versions of the same object type or due to different conditions of conservation (wear, cleanliness, etc.)

\paragraph{Learning from instructional videos.} 
Learning from instructional videos has been recently addressed in the context of problems such as action discovery~\cite{alayrac:hal-01580630,richard18actionsets,sener18unsupervised,Kukleva_2019_CVPR,elhamifar20} and localization~\cite{Huang2016connectionist,richard17weakly,youcook2,Zhukov_2019_CVPR}, modeling of object states~\cite{alayrac16objectstates} and visual reference resolution~\cite{huang17unsupervised,Huang_2018_CVPR}.
A particular attention has been drawn to the {\em narrated} instructional videos, \ie instructional videos, accompanied by comments in the form of natural language. It was shown, that narrations in such videos generally describe the visual content and can be used as an additional guidance for learning the visual models~\cite{Malmaud15what,alayrac:hal-01580630,Zhukov_2019_CVPR}, or for joint modeling of video and language~\cite{Miech_2019_howto100m,miech_e2e,Zhu20}. In this work, we make use of narration for object grounding in 3D, assuming that object mentions in narration coincide with their appearance on the screen. 

Related to our work, Damen et al.,~\cite{dima2014youdo} consider the problem of task-relevant object discovery from videos. Unlike~\cite{dima2014youdo} we don't aim for a discovery of objects, and only attempt to localize objects in the 3D scene. On the other hand we consider a more general setup: our videos depict different object instances in uncontrolled scenarios and are sourced from a mix of egocentric and third-person cameras.

\paragraph{Visual language grounding.}
 Visual language grounding requires associations between the text the most relevant parts of an image. A standard approach to this task involves learning a joint embedding of language and visual data and comparing the similarity between the text and image regions in the joint embedding space~\cite{Karpathy14deepFragment,Wang_2016_CVPR_embedding,10.1007/978-3-319-46484-8_42,8268651,plummer2018eccv}. Most works focus on learning to ground language in images in supervised settings, where the regions of images, relevant to the input text are provided at training time. On the contrary,~\cite{Zhao_2018_CVPR,Engilberge_2018_CVPR,Xiao_2017_CVPR} consider weakly-supervised scenario where only the correspondences between text and images are provided during training, while the exact regions, where the text is grounded, are unknown.

Closer to our work, Russell \etal~\cite{russel2013acm} attempt to localize text from Wikipedia articles in a 3D scene. This work focuses on museum scenes and the objects, such as paintings and frescos. This is done in two steps. First, an object caption is used as a query to search for related images with Google image search. Retrieved images are then registered to the 3D scene and their extents are used as a clue for object localization in 3D. The authors of this work leverage the {\em photographer's bias}, \ie the fact that objects of interest are usually perfectly framed by the extents of the image. Our solution for the 3D text grounding is based on a similar observation. However, we focus on instructional videos and unlike~\cite{russel2013acm}, we use the same video data both for reconstruction and for language grounding, and don't rely on query expansion via external image search.

\section{3D reconstruction of instructional videos}
\begin{figure*}[t]
\centering
\captionsetup[subfigure]{labelformat=empty}
\begin{subfigure}[b]{0.00001\textwidth}
\caption{\label{fig:overview_a}}
\end{subfigure}
\begin{subfigure}[b]{0.00001\textwidth}
\caption{\label{fig:overview_b}}
\end{subfigure}
\begin{subfigure}[b]{0.00001\textwidth}
\caption{\label{fig:overview_c}}
\end{subfigure}
\begin{subfigure}[b]{0.00001\textwidth}
\caption{\label{fig:overview_d}}
\end{subfigure}
\includegraphics[width=0.9\textwidth]{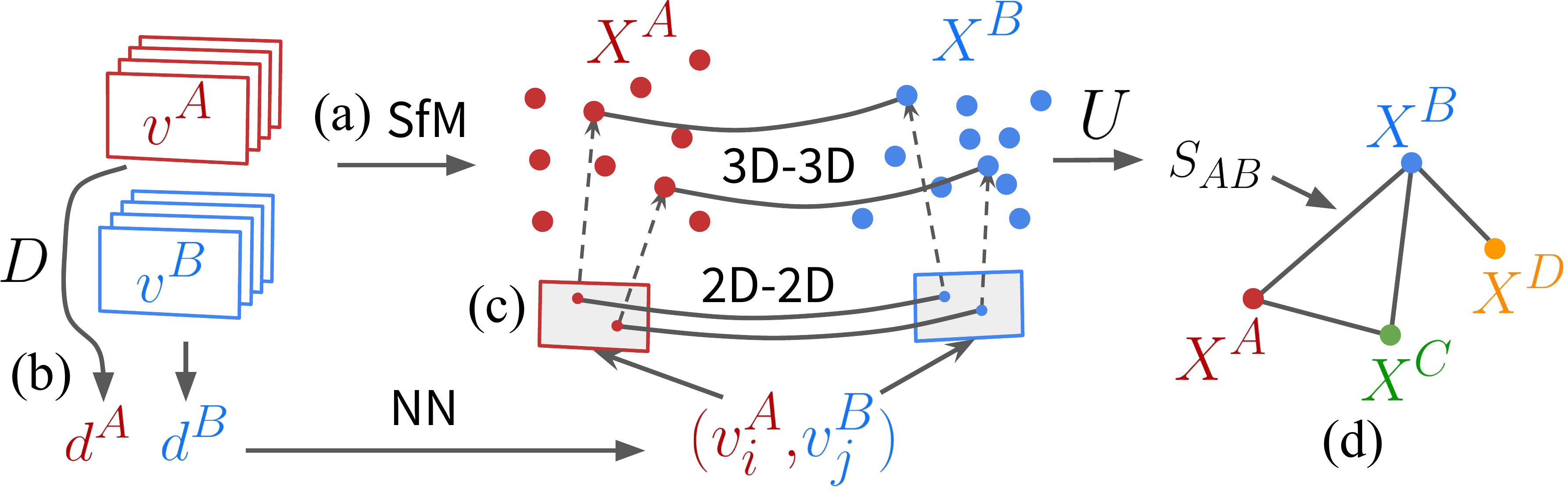}
 \vspace*{-0.4cm}
\caption{\textbf{Proposed approach for reconstructing instructional videos in 3D.} (a) Given two videos, {\color{BrickRed} $v^A$} and {\color{NavyBlue} $v^B$}, each is independently reconstructed, producing the 3D models {\color{BrickRed} $X^A$} and {\color{NavyBlue} $X^B$}. (b) Furthemore, image-level descriptors {\color{BrickRed} $d^A$} and {\color{NavyBlue} $d^B$} are extracted with $D$, and used to obtain candidate pairs of frames $({\color{BrickRed}v^A_i}, {\color{NavyBlue}v^B_j})$. (c) Then, 2D-2D  correspondences between these pairs of frames are obtained by matching local features with global constraints imposed by dense flow estimation, and 3D-3D correspondences are established from the 2D-2D correspondences by leveraging the individually reconstructed 3D models, {\color{BrickRed} $X^A$} and {\color{NavyBlue} $X^B$}, and the camera parameters (extrinsics and intrinsics) for {\color{BrickRed}$v^A_i$} and {\color{NavyBlue}$v^B_j$}. (d) The aggregated 3D-3D correspondences from several image pairs are passed to the solver $U$ that robustly estimates a similarity transformation. Finally, a graph is constructed where each node corresponds to a 3D model, and each edge corresponds to a successful alignment between a pair of point-clouds. By traversing this graph, and composing the transformations along a given path, one can compute the similarities between any pair of connected nodes.
}
 \vspace*{-0.5cm}
\label{fig:overview}
\end{figure*}

In this section we present our approach for reconstructing narrated instructional videos in 3D. 
In order to address the limitations of the standard instance-level 3D reconstruction pipeline (i.e. limited robustness to appearance changes and large computational complexity) we propose an approach based on the \emph{divide and conquer} strategy, where each video is reconstructed independently and then the resulting 3D models are combined into a 3D alignment graph that allows to obtain a final consistent model. In this way we have two advantages over the joint reconstruction approach. First, only a few correspondences are needed \emph{across} different videos, in order to estimate the similarity transformation that aligns the 3D models reconstructed from each video. Second, the computational complexity of the 3D reconstruction is reduced from $O(n^2N^2)$ to $O(N^2)$, where $N$ is the number of frames sampled from each video, and $n$ is the number of videos. In the following sections, we describe in detail our approach for reconstructing instructional videos in 3D. An overview of the approach is presented in Fig.~\ref{fig:overview}.

\subsection{Per-video reconstruction \label{sec:pairvideorec}}
In order to avoid the large computational complexity and matching challenges of joint 3D reconstruction, we reconstruct each video independently. For each video, we proceed in the following way. First, $N$ video frames $\{v_i\}_{i=1,\dots,N}$ are randomly sampled. Then, CNN-based local features $(f_i,p_i) = L(v_i)$ are extracted for every frame using the feature transform $L$, where $f_i\in\mathbb{R}^{d_L\times N_L}$ is the set of  $d_L$-dimensional feature descriptors, $p_i\in\mathbb{R}^{2\times N_L}$ is the set of 2D key-point positions, and $N_L$ is the number of extracted local features. Then, exhaustive matching is performed between the features of every video frame $v_i$. Only matches that are mutually nearest-neighbors are retained. Finally, these matches are input into an SfM pipeline to produce a 3D model represented as a cloud of 3D points $X=\{(X_i,Y_i,Z_i)\}_{i=1,\dots, N_p}$ for the given set of video frames\footnote{In some cases, several 3D models will be produced due to the fact that videos might have edits, and show completely different scenes at different times. Small models are discarded.}. An illustration of this approach is presented in Fig.~\ref{fig:overview_a}.

\subsection{Formation of frame-pairs across videos \label{sec:framepairs}}
Once every video is independently reconstructed, the goal is to align them to a common frame of reference, which involves estimating a 3D similarity transformation. Then, given a pair of videos $(v^A,v^B)$, the first step involves estimating a candidate set of matching video frames in the form of $M=\{(v^A_i, v^B_j)\}$, where $i$ and $j$ are the frame indices of matching frames of $v^A$ and $v^B$ respectively. In order to achieve this, we follow the approach that is typically used for image retrieval. This consists of, first, extracting image-level descriptors $d^A_i = D(v^A_i)$ and $d^B_j = D(v^B_j)$ using the image-level description function $D$ for every frame $v^A_i$ in $v^A$ and $v^B_j$ in $v^B$, where $d^A_i, d^B_j \in \mathbb{R}^{d_I}$ are $d_I$-dimensional descriptors. Then, the top $N_M$ matches between videos $v^A$ and $v^B$ are selected by nearest-neighbor search between the sets of image level descriptors $\{d^A_i\}$ and $\{d^B_j\}$. An illustration if this approach is presented in Fig.~\ref{fig:overview_b}.

\subsection{Establishing 2D correspondences between videos\label{sec:robustmatches}}
Once candidate matching frames $\{(v^A_i,v^B_j)\}$ between a pair of videos $(v^A,v^B)$ are established, the next step consists of robustly estimating 2D-2D correspondences between them.
In order to handle the large appearance differences that may be present in these two frames, we propose to use a combination of learnt local features and dense flow to robustly obtain matches that are simultaneously well localized and compliant with a global transformation. Given a pair of frames $(v^A_i,v^B_j)$, mutual matches between the sets of CNN-based local features $(f^A_i,p^A_i)$ and $(f^B_j, p^B_j)$ are first computed. Using local features allows for a very precise localization of the matches, but does not enforce any global consistency among different matches. In practice, this may lead to a high fraction of incorrect correspondences that may cause the subsequent alignment estimation step to fail. In order to reduce the fraction of incorrect correspondences while retaining the correct correspondences, we filter the matches produced by local features using a dense transformation field $\mathcal{T}^{AB}_{ij} = G(v^A_i,v^B_j)$ obtained from a dense flow model $G$, where $\mathcal{T}^{AB}_{ij}: \mathbb{R}^2\rightarrow \mathbb{R}^2$ is a 2D-2D mapping between the pixels of $v^A_i$ and those of $v^B_j$, and $G$ is a CNN-based dense flow estimation model. Only matches between local features that are consistent with the dense flow mapping $\mathcal{T}^{AB}_{ij}$ up to certain tolerance threshold $t$ are retained. An illustration if this approach is presented in Fig.~\ref{fig:overview_c}.

\subsection{Estimating 3D transformation between videos}
Once 2D-2D correspondences between frames of different videos are established, these can be used to, first, establish 3D-3D correspondences, and, second, determine a 3D similarity transformation between these videos.

Given a pair of videos $(v^A,v^B)$, 2D-2D correspondences are first aggregated from all its matching frames $\{(v^A_i,v^B_j)\}$. Then, these 2D-2D correspondences are converted to 3D-3D correspondences by leveraging the camera parameters (intrinsic and extrinsic) together with the 3D models $X^A$ and $X^B$ obtained from their individual 3D reconstructions (cf. Sec.~\ref{sec:pairvideorec}). Then, the set of 3D-3D correspondences is input to a solver $U$ that robustly estimates the similarity transformation $S_{AB}$ between 3D models $X^A$ and $X^B$. However, some models may be difficult to align directly due to very large viewpoint changes, which may cause the matching method presented in Sec.~\ref{sec:robustmatches} to fail. To overcome this issue, we propose to construct a \emph{3D alignment graph} where every node represents a 3D model $X^m$ and every edge represents a successful alignment between the 3D models of the two nodes that constitute the edge. Then, the alignment between any two connected models in the graph can be estimated by computing the shortest path~\cite{dijkstra1959note} and then composing the pairwise transformations along the path. Note that by traversing the graph it is possible to align 3D models across large changes of viewpoint or appearance which would not be possible by direct matching. In order to register the models to a common reference frame, any node can be selected as reference and the similarities to every other node computed by traversing the graph. An illustration of this approach is presented in Fig.~\ref{fig:overview_d}. Appendix~\ref{sec:sup_alignment} provides additional details.

\subsection{Application to keypoint transfer}
In this section we describe how the proposed 3D alignment graph can be used to perform keypoint transfer among models. 
Because manual annotation is costly,  annotation transfer has been used in the past as a way to obtain annotation on new examples in an automatic way~\cite{guillaumin2014imagenet,wang2019learning,kim2017fcss}. In our case, we consider the problem of transferring annotations across several instructional videos showing the same object model. In particular, consider that we have a source video $v^s$, and a set of target videos $\{v^{t_1},v^{t_2},\dots,v^{t_{N_t}}\}$, where $N_t$ is the number of target videos. Consider as well that we are interested in a set of $N_k$ keypoints. Then, we manually annotate these $N_k$ keypoints on several frames of the source video $v^s$ and triangulate their 3D positions $K^s\in \mathbb{R}^{3\times N_k}$ in the source video reference frame, where the column of $K^s$ correspond to the 3D coordinates of each different keypoint.
Then, we can apply the similarity transformation $S_{s,t_i}$ between the source video $v^s$ and each target video $v^{t_i}$ (obtained by traversing the 3D alignment graph) to transfer the keypoints $K^s$ to each target video $v^{t_j}$:
\begin{equation}
    K^{t_i} = S_{s,t_i} K^s
\end{equation}

Finally, these transferred 3D keypoint annotations can be projected to any 2D frame $v^{t_i}_j$ of the target video $v^{t_i}$ by using the estimated projection matrix $P^{t_i}_j$ (obtained from the product of the intrinsic and extrinsic camera parameters).

\section{Text grounding in 3D\label{sec:grounding}}
Given a 3D scene reconstruction, our goal is to obtain 3D text grounding, i.e., to identify the parts of the 3D model corresponding to a given text query. 
We follow~\cite{Alayrac15Unsupervised,Miech_2019_howto100m} and assume video narrations to be approximately synchronized with video frames of corresponding instructional videos and to describe their visual content. 
We consider alternative methods for localizing narrations in 2D frames and then backproject such location onto the 3D model.
While the individual backprojected 3D positions may be noisy, e.g. because of the misalignment of the narration and the video, we obtain accurate positions by accumulating evidence in 3D from multiple videos.

\begin{figure*}[t]
\begin{center}
   \includegraphics[width=\linewidth]{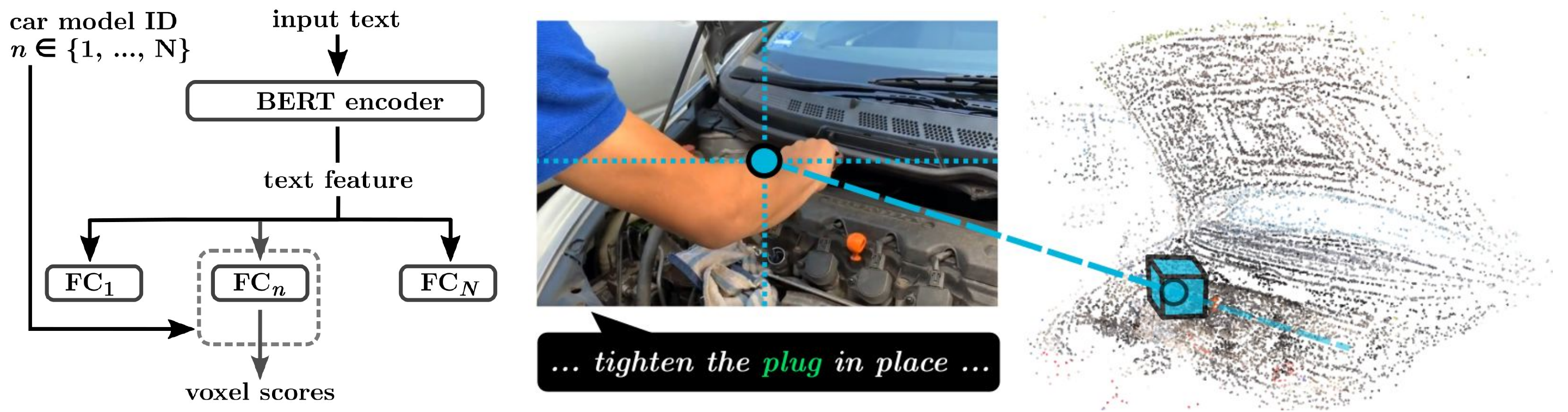}
   \vspace*{-0.6cm}
\caption{{\bf Left.} Illustration of our 3D text grounding model. We train a separate model for each car make. All models share the same text encoder (BERT) and differ in the last grounding layer. The grounding layer is a single fully connected layer with the output dimension equal to the number of voxels. Given the input text, the model produces the confidence scores for all voxels in the 3D model. \textbf{Middle and Right.} Illustration of our training procedure. Given a word sequence from subtitles (bottom), and a 2D point in the frame coordinates (\textcolor{cfblue}{blue} dot), we backproject the point onto the surface of the 3D model and find the corresponding voxel (\textcolor{cfblue}{blue} cube). The pair (text, voxel) is then used as a training pair. In this example, we use the center of the frame as the approximate supervision to ground the text narration on the 3D model.
}
\vspace*{-0.9cm}
\label{fig:ground_method}
\end{center}
\end{figure*}

Our approach is summarized in Fig.~\ref{fig:ground_method}. We first employ a text encoder model $B$, that produces a $d-$dimensional text feature descriptor $f_s=B(s)$ from a given input text string $s$. We then use a classifier $C$ to predict the distribution of scores $c=C(f_s)\in\mathbb{R}^n$ over a set of $N_V$ voxels that discretize the 3D model space, for the given text features $f_s$. In the following, we describe our approach for training such classifier and text encoder in an unsupervised way from narrated instructional videos.

\medskip
\subsection{Generation of training data}
We propose to automatically generate training samples for learning text grounding in 3D scenes.
Our objective is to obtain training pairs $(s_i, k_i)$ formed by the text $s_i$ and the corresponding location in the scene represented by the 3D voxel label $k_i\in[0,N_V]$. Towards this goal we first split an instructional video $v$ and its corresponding transcribed narration $s$ into a set of temporal segments $\{(v_i,s_i)\}_{i=1,\dots, n}$. To obtain $k_i$ for each segment, we first select a 2D position $p_i=(x_i,y_i)$ in video $v_i$, and then use the camera parameters (intrinsics and extrinsics) together with the reconstructed 3D model to obtain the corresponding 3D point $P_i$ of the scene.
We build a pair $(s_i, k_i)$ by choosing a 3D voxel $k_i$ containing $P_i$.

To obtain 2D image locations $p_i$, we ground narrations $s_i$ in video frames by following three alternative approaches.
\smallskip\\
{\bfseries Center of frame.} The first approach is to assign $p_i$ to the frame center. This simple strategy turns out to be effective in instructional videos where narrated objects and actions are typically centered in the frame by the cameraman.  
\smallskip\\
{\bfseries Hand detector.}
Our second approach follows the assumption that narrated objects are being manipulated by the user. We hence assign $p_i=H(v_i)$, where $H$ is a position of a hand in the video obtained with a hand detector~\cite{Shan_2020_CVPR}. 
\smallskip\\
{\bfseries MIL-NCE.}
Our third approach builds on a pre-trained model for local text-video similarity. We use a joint text-video embedding model trained on a large set of narrated instructional videos with MIL-NCE~\cite{miech_e2e}. This model projects video and text to a joint embedding space, which allows us to compute similarity between a local spatio-temporal video region and a text fragment. To ground the text narration within the frame, we select a position in a feature map with the highest text-video similarity score.

In all three cases described above, the selected 2D locations $p_i$ within frames are reprojected to the surface of the reconstructed 3D model to obtain $P_i$ and corresponding voxels $k_i$.

\subsection{Learning 3D grounding model}

Given a set of training pairs with narrations and corresponding voxels $\{(s_i, k_i)\}_{i=1,\dots, n}$, the classifier $C$ is trained to predict the correct voxel label $k_i$ given input text $s_i$. We train $C$ by minimizing the cross-entropy loss over all training pairs:
\begin{equation}
    \label{eq:3d_loss}
    -\sum\limits_{i=1}^n\log\left(C(B(s_i))[k_i] \right),
\end{equation}
where $C(B(X))[k]$ is the score of voxel $k$, given input text $X$, according to our model $C$. This loss is minimized when $C$ assigns the maximum score of one to the correct voxel.

In practice, we are interested in training language grounding models for different 3D reconstructions. For example, for car maintenance videos, we build a separate 3D reconstruction for each specific car model, such as Ford Focus or Honda Civic, as the 3D shape of the car and the engine can be substantially different. However, while the classifier $C$ is different for all 3D reconstructions, we choose to share the same text encoder $B$ for all reconstructions. This allows us to benefit from all available text data, when training the text representation module. The entire model trained in a multi-task setting is shown in Fig.~\ref{fig:ground_method}.
Additional implementation details are provided in Sec.~\ref{sec:results}.

\begin{figure}[t]
\centering
\includegraphics[width=\columnwidth]{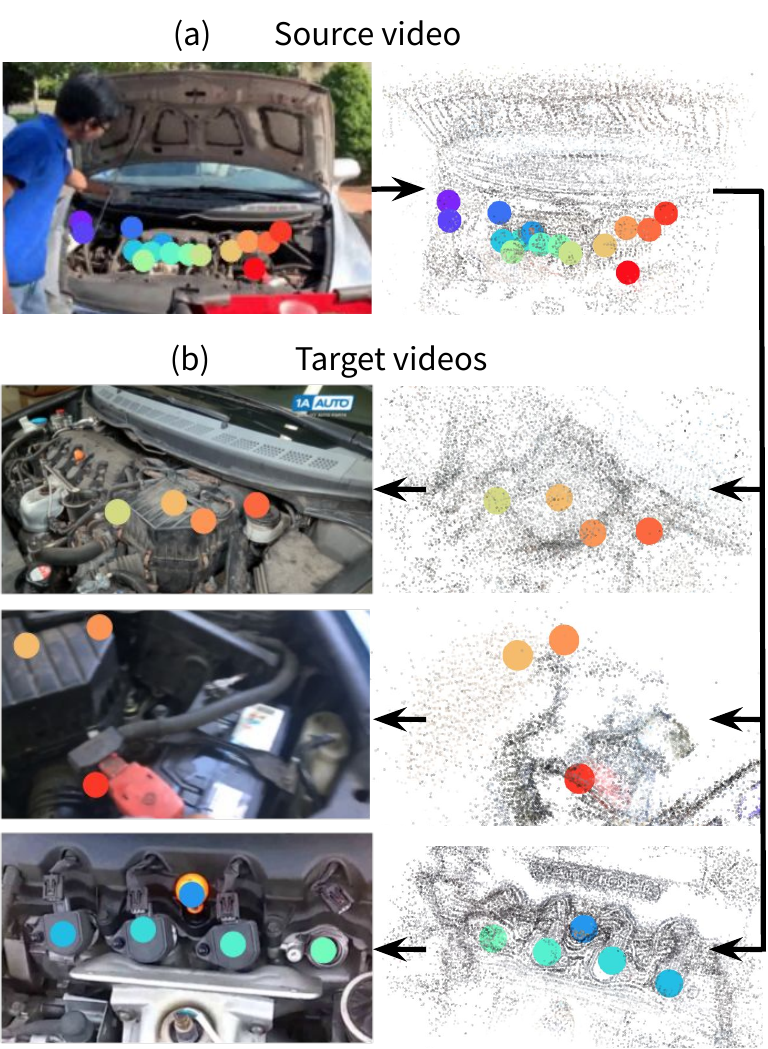}
 \vspace*{-0.6cm}
\caption{\textbf{Example of keypoint transfer results.} (a) 2D keypoints are annotated in several frames of the source video and triangulated to 3D keypoints in the 3D reference frame of source video. (b) These source keypoints are automatically transferred to the 3D reference frames of several target videos using the proposed alignment graph. The 3D keypoints can be finally projected to 2D keypoints for every registered frame in the target videos.}
 \vspace*{-0.7cm}
\label{fig:kp_transfer}
\end{figure}

\section{Experimental results
\label{sec:results}}

To evaluate our proposed method for 3D reconstruction and 3D grounding, we collect a dataset of narrated instructional videos related to car maintenance tasks. The tasks we consider are the following: replacing the air filter, changing the spark plugs, replacing the car battery, measuring the oil level, adding coolant fluid and checking the break fluid. The car models we consider are: Honda Civic (8th gen), Honda Accord (7th gen), Ford Focus (Mk3), Ford Explorer (5th gen), Toyota Corolla (E140) and Toyota Prius (XW20). We select these car models as they are common in North America. We also focus our data collection on English-spoken videos. We have collected a total of 174 videos, with the number of videos per car model between 23 and 36. The number of videos per task varies between 14 (measure oil level) and 53 (change spark plugs). 

In the following we first describe implementation details and then present results obtained for the keypoint transfer and 3D text grounding tasks. Our dataset, pre-trained models and code will become publicly available. 

\paragraph{Implementation details.} We use R2D2~\cite{r2d2} learnt local features as the feature extraction function $L$ and GLU-Net~\cite{truong2020glu} as the learnt dense flow estimation function $G$. The image level description-function $D$ is done using a NetVLAD~\cite{arandjelovic2016netvlad} model with a VGG16 backbone~\cite{simonyan2014very}. The solver $U$ is composed by a RANSAC loop~\cite{fischler1981random} that performs a transformation fitting using the least-squares method proposed by Umeyama~\cite{umeyama1991least}.

For the text grounding, we use a pre-trained BERT model as the text encoder $B$. The output dimension $d$ of the text encoder is set to 1024. A linear classifier is used for the grounding layer $C$, where the output dimensionality corresponds to the number $N_v$ of voxels. The text enconder $B$ is finetuned together with the grounding classifier $C$ in an end-to-end manner. For the hand detector $H$, we use~\cite{Shan_2020_CVPR}. When using MIL-NCE~\cite{miech_e2e} we obtain local similarity scores for any location of the frame by removing the last average pooling layer from the S3D backbone of the visual branch of the model and obtain a features map of size $14\times8$ for each selected frame of the video.

We define a set of voxels as follows. First, consider a rectangular region, that encompasses the 3D model. This region is divided into 20 equal parts by each axis, resulting in a set of 8000 voxels. 
Since the shape of a car engine is relatively flat, we may expect that most of these voxels don't intersect with any parts of the model. In order to reduce the total number of voxels, and only retain the voxels, that intersect with objects of interest in the 3D scene, we select top $N_v$ voxels, that contain most of the training points. In practice $N_v$ is set to 500.

\subsection{Keypoint transfer}
For evaluating the task of keypoint transfer we annotate multiple frames for each Honda Civic video. We select one video as source ($v^s$), and the rest as target ($\{v^{t_i}\}_{i=1,\dots,N_t}$). Then, the annotations are used to triangulate the 3D keypoint positions $K^s$ for the source video and for each of the target videos $K^t_i$, where $i=1,\dots,N_t$. Then, we estimate a ground-truth similarity transformation between the source video and the target videos by fitting a similarity transformation using the solver $U$ on the triangulated keypoints:
    
\begin{equation}
    S^{GT}_{s,t_i} = U(K^s, K^{t_i}).
\end{equation}

\begin{figure}[t]
\centering
\includegraphics[width=\columnwidth]{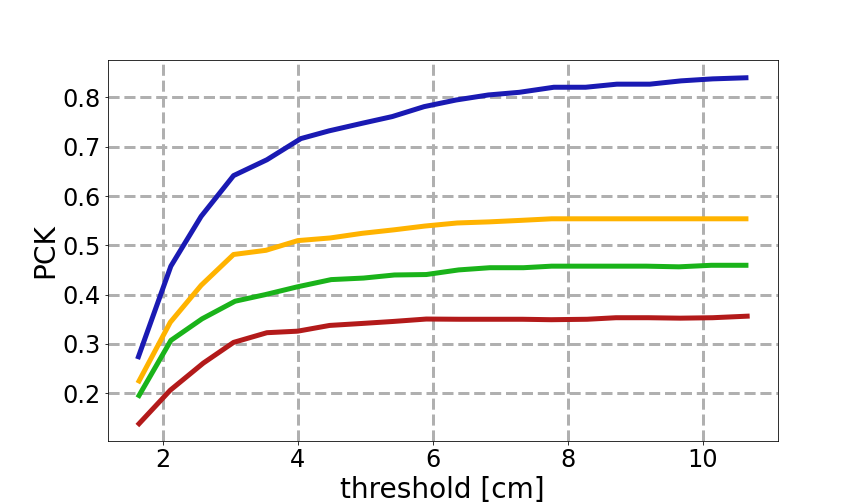}
{\small
\begin{tabular}{@{}llll@{}}
\includegraphics[width=0.5cm]{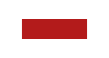} & \cite{r2d2} &
\includegraphics[width=0.5cm]{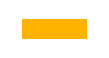}  & Ours (direct matching)
\\
\includegraphics[width=0.5cm]{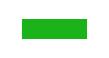}  & \cite{truong2020glu} &   \includegraphics[width=0.5cm]{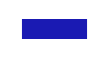} & Ours (3D alignment graph) \end{tabular}
}
 \vspace*{-0.2cm}
\caption{\textbf{Quantitative evaluation of keypoint transfer.} We plot the mean PCK curve over a set of 89 video pairs. The proposed 3D alignment graph correctly aligns 80\% of keypoints within a 8cm tolerance.}
 \vspace*{-0.6cm}
\label{fig:transfer_pck}
\end{figure}

Finally, we can assess the quality of the similarity transformations $S_{s,t_i}$ that we estimated using the proposed 3D alignment graph by computing the percentage of correctly transferred keypoints (PCK) in 3D coordinates.
For evaluation, we use 89 pairs of videos for the Honda Civic car model, where the source and target keypoints $K^s$ and $K^t$ have been verified to have consistent relative distances and angles (a consistent overall 3D shape). Our results are presented in Fig.~\ref{fig:transfer_pck}, where we plot the mean PCK over these 89 evaluation pairs vs. a varying tolerance threshold calibrated in centimeters. Results show that our proposed matching approach using a combination of R2D2 learnt local features and GLU-Net dense flow is superior to these two methods employed individually. Nevertheless, some video pairs are very difficult to match directly due to extreme viewpoint changes and our direct matching approach can only align about 50\% of the keypoints for a 10cm tolerance value. In this cases the 3D alignment graph can be used to compute the similarity transformation via other intermediate videos. Results show that by using the 3D alignment graph, 80\% of keypoints are correctly localized within a tolerance of 8cm. Examples of keypoint transfer using the 3D alignment graph are shown in Fig.~\ref{fig:kp_transfer} and, thanks to the 3D alignment graph, demonstrate transfer across an impressive range of  viewpoints and appearance variation across the different car instances. 

\subsection{Text grounding in 3D\label{sec:5.2}}

\vspace*{-1mm}
\paragraph{Evaluation setup.}
To demonstrate the effectiveness of 3D text grounding, we select six different objects: \textit{air filter}, \textit{brake fluid reservoir}, \textit{negative battery cable}, \textit{positive battery cable}, \textit{dipstick} and \textit{spark plugs}, and annotate their positions within each 3D model. Each object is annotated with a single 3D point. We manually parse the WikiHow.com car repair articles and select a set of textual instructions, related to each object. These instructions are object-centric, each refering to a specific location within the engine, where one of the objects is located. Examples of such instructions include "remove the dipstick" (\textit{dipstick}), "use a wrench to loosen the positive cable clamp and take the cable off of the terminal" (\textit{positive battery cable}) and "lift the filter out of the housing" \textit{(air filter}). At test time, we use these instructions as queries, with the goal to ground each query  to the correct 3D location, defined by the corresponding point of interest. Since our model returns the scores over a set of voxels, instead of a single 3D point, we define the grounding by the model as the center of a highest scoring voxel.

\vspace*{-2mm}
\paragraph{Evaluation metric.}
We measure the performance of the grounding by the PCK score. This score is computed as follows. For each car model and for each query, we compute the distance between the predicted 3D point and the ground truth 3D point that corresponds to the object within the query. Then for each threshold value from 1cm to 100cm we compute PCK as the number of queries grounded correctly (\ie the distance between the predicted point and the ground truth point is below the given threshold value), divided by a total amount of queries for all models. 

\begin{figure}[t]
   \includegraphics[width=\columnwidth]{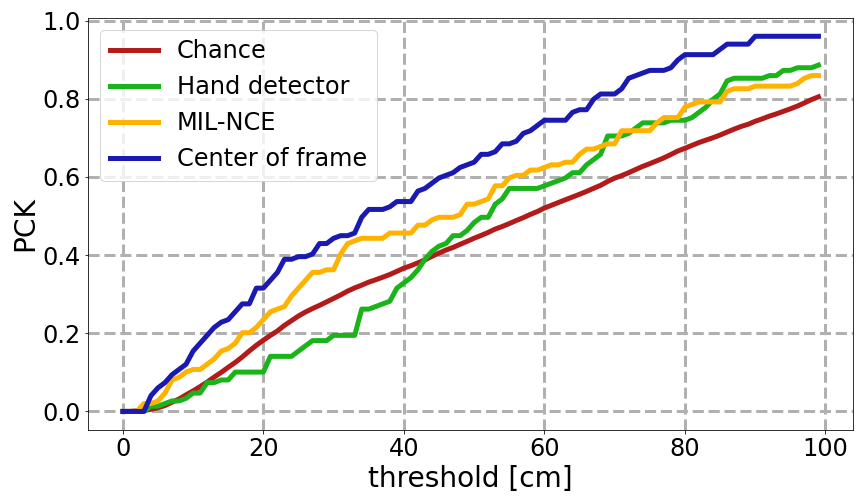}
    \vspace*{-0.8cm}
\caption{{\bf Quantitative evaluation of 3D text grounding.} We plot PCK curve over a set of 6 different car models and 6 different objects of interest. For each object we consider a set of text queries, related to the corresponding location in the 3D model, and predict the 3D location of the object based only on the the language query. We compare our model trained with different types of supervision against a Chance baseline, where each query is grounded in a random point on the 3D model.
}
 \vspace*{-0.6cm}
\label{fig:ground_pck_4.2}
\end{figure}

\begin{figure}[t]
\begin{center}
   \includegraphics[width=\columnwidth]{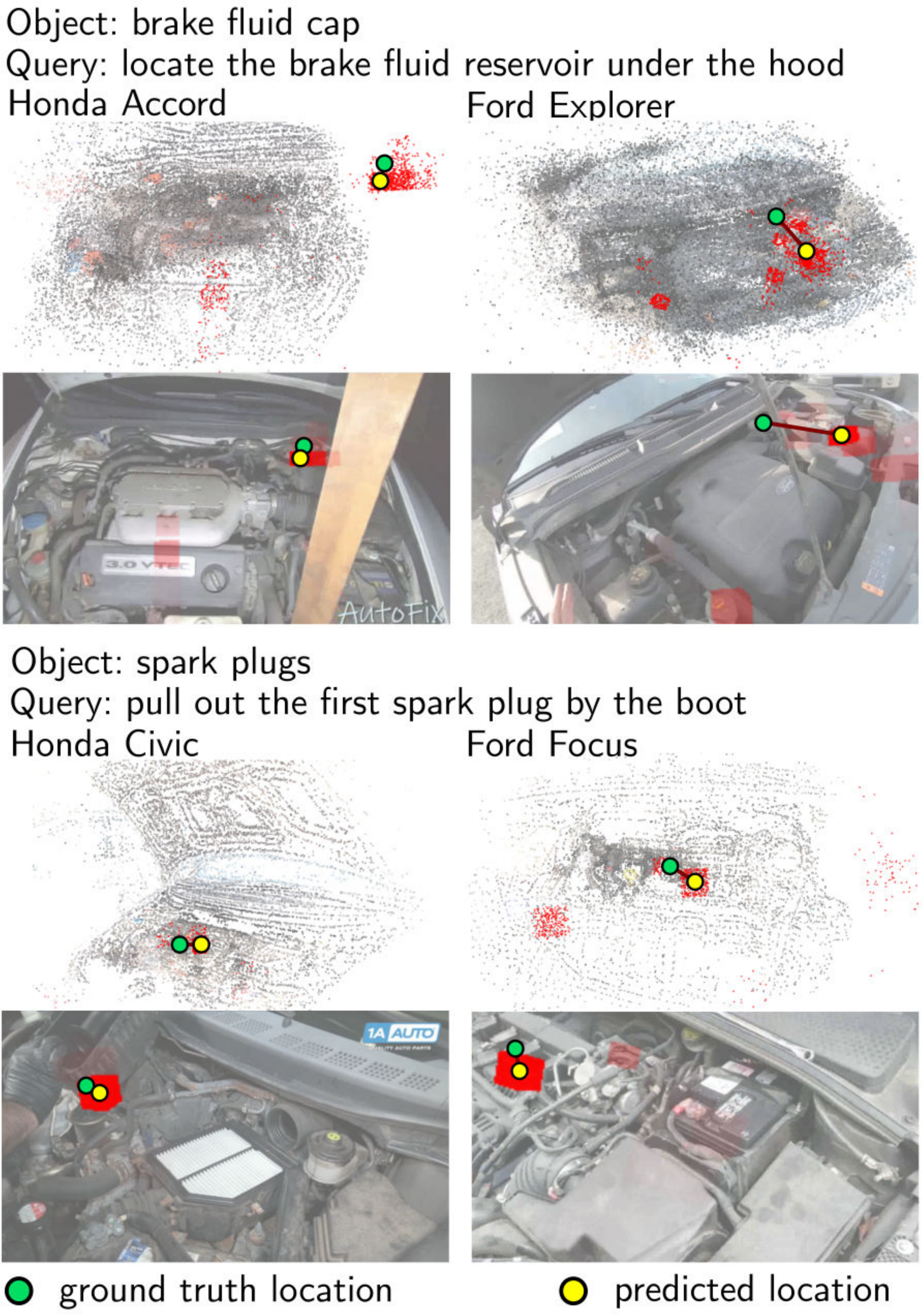}
\caption{{\bf Examples of 3D text grounding with our model.} Our model predicts a heatmap over a set of voxels with higher values assigned to locations that correspond to the query text. For clarity, we visualize the predicted voxels in random frames from the videos. The confidence that a given text query corresponds to a given location is shown in \textcolor{red}{red} with higher opacity corresponding to higher confidence. The ground truth 3D location is shown by a \textcolor{gtgreen}{green} dot. Please see additional results in Fig.~\ref{fig:text_ground_sup}.
}
\vspace*{-0.35cm}
\label{fig:qualitative_4.2}
\end{center}
\end{figure}

\vspace*{-2mm}
\paragraph{Results and discussion.}
We evaluate our models, trained with three different types of supervision, as described in Sec.~\ref{sec:grounding}, and compare their performance against a chance baseline (``Chance"), which implies grounding each query to a random point in the 3D model. Results are shown in Fig.~\ref{fig:ground_pck_4.2}.
Interestingly, the best results are provided by the center of frame method (``Center of frame")  confirming the previously observed photographer bias~\cite{russel2013acm}. 
Supervision from the Hand detector (``Hand detector") suffers from a large amount of noise when the hands are visible in the video but are not near the object. The hand detector also provides less supervision than other methods, since hands may not appear within the frame at the time when the object is mentioned by the narration. Finally, in many cases, the appearance of hands in the videos coincide with an object mention in narration, when an action over the object is being performed. However, an action often involves using tools, such as a wrench or a screwdriver. Therefore, the position of the hand is often shifted with respect to the object, contributing to the noise in the supervision.
We found MIL-NCE feature (``MIL-NCE") not well suited for the task. While, in some cases, this feature is able to provide good localization, it is usually very coarse, as MIL-NCE is not designed to accurately localize queries within the image. 
Results for different object classes are shown in Tab.~\ref{table:pck_classes}. These results indicate that our model (here the best performing variant ``Center of frame" is used) consistently outperforms the chance baseline for 5 object classes out of 6, and only fails on the dipstick object.
Fig.~\ref{fig:qualitative_4.2} demonstrates qualitative results for different queries and different car models. Additional qualitative results are provided in Appendix~\ref{sec:sup_grounding}.

\begin{table}[t]
    \centering
    \begin{tabular}{@{}lcc@{}}
      \toprule
      Object & Chance & Our method \\
      \midrule
		Air filter & 0.26 & \textbf{0.55} \\
		Brake fluid reservoir & 0.19 & \textbf{0.23} \\
		Negative battery cable & 0.07 & \textbf{0.42} \\
		Positive battery cable & 0.05 & \textbf{0.20} \\
		Dipstick & \textbf{0.40} & 0.35 \\
		Spark plugs & 0.54 & \textbf{0.69} \\
      \midrule
      Average & 0.29 & \textbf{0.41} \\
      \bottomrule
    \end{tabular}
     \vspace*{-0.0cm}
 \caption{PCK grouped by object classes. We compare our model, trained on centers of frames, against the Chance baseline. Here the PCK error threshold is set to 30cm.}
       \vspace*{-0.0cm}
      \label{table:pck_classes}
\end{table}

\section{Conclusion}
We present a method for 3D reconstruction of instructional videos and localizing the associated narrations in 3D. Our method is resistant to the differences in appearance of objects depicted in the videos and computationally efficient. Building on our 3D reconstruction, we propose an unsupervised approach to 3D language grounding.
Our work opens-up the possibility of providing visual hints, based on textual instructions for augmented reality applications.  
\paragraph{Acknowledgements.} This work was supported in part by 
the MSR-Inria joint lab, 
the Louis Vuitton ENS Chair on Artificial Intelligence, 
the European Regional Development Fund under the project IMPACT (reg. no. CZ.02.1.01/0.0/0.0/15003/0000468),
and the French government under management of Agence Nationale
de la Recherche as part of the “Investissements d’avenir” program, reference
ANR-19-P3IA-0001 (PRAIRIE 3IA Institute). 
This work was granted access to the HPC
resources of IDRIS under the allocation 20XX-AD011011329.

{\small
\bibliographystyle{ieee_fullname}
\bibliography{biblio}
}

\setcounter{section}{0}
\renewcommand{\thesection}{\Alph{section}}
\clearpage
\appendix
\onecolumn
\section*{Appendix}

This appendix presents additional
results of our method. In Sec.~\ref{sec:sup_alignment}, we provide additional qualitative results demonstrating the benefits of using the 3D alignment graph.
In Sec.~\ref{sec:sup_grounding} we provide additional qualitative results for text grounding in 3D. 

\section{Estimating 3D transformation between videos\label{sec:sup_alignment}}
\begin{figure}[b!]
\includegraphics[width=\linewidth]{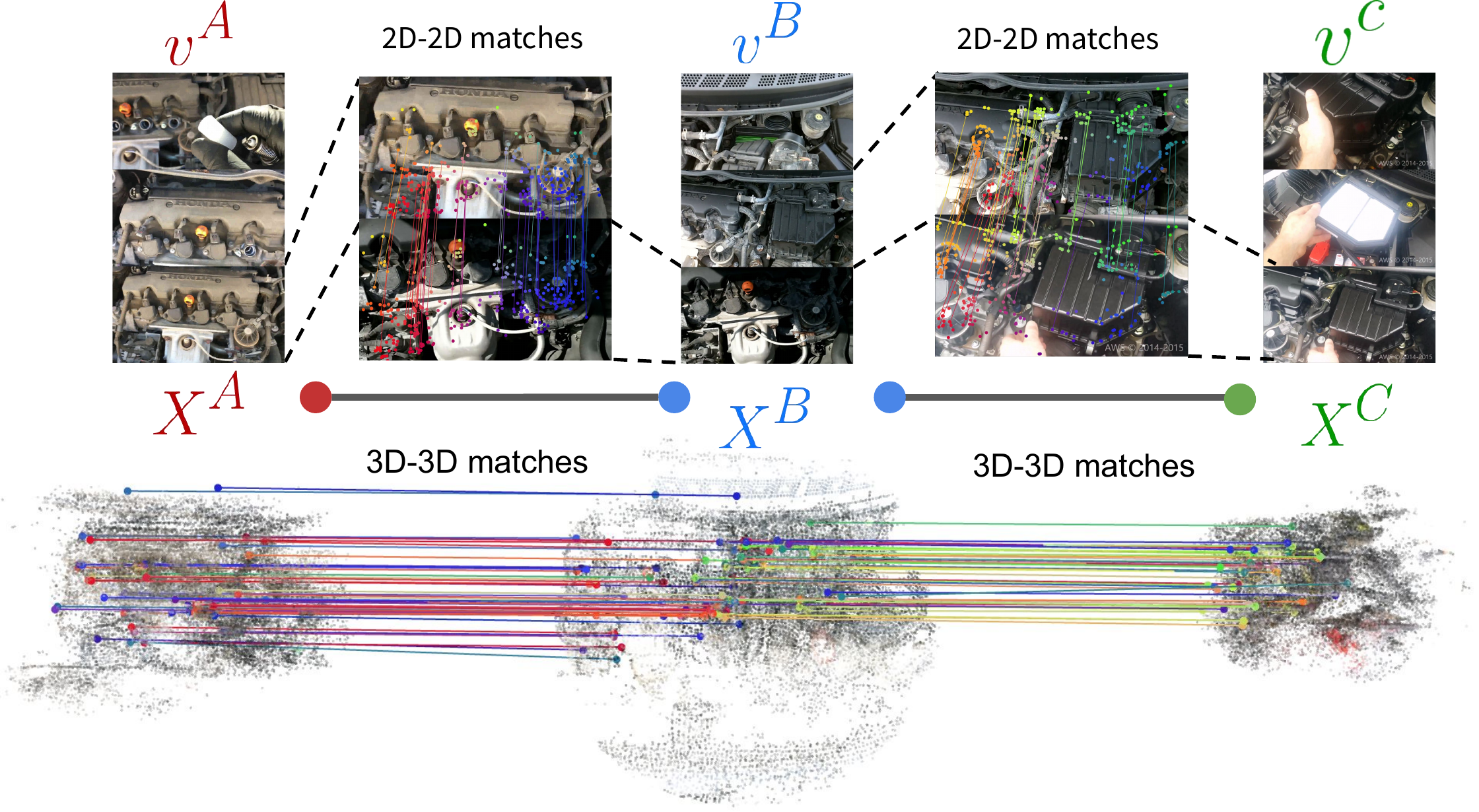}
\caption{{\bf Illustration of the proposed alignment method and the 3D alignment graph.} Given two videos $v^A$ and $v^B$, pairs of matching frames are found using global image descriptors, and are then matched using our approach to obtain 2D-2D matches (top). These 2D-2D matches are then translated into 3D-3D matches between the 3D point clouds $X^A$ and $X^B$ corresponding to $v^A$ and $v^B$, respectively (bottom). These 3D-3D matches are used to estimate the similarity transformation $S_{AB}$ between $X^A$ and $X^B$, using the function $U$. Furthermore, the constructed 3D alignment graph can be used to align two videos $v^A$ and $v^C$ which have little overlap (i.e. they focus on different parts of the object), by traversing the edges of the graph using one or more intermediate videos and 3D models ($v^B$ and $X^B$ in this example). 
}
\label{fig:alignment}
\end{figure}

In this section we further illustrate details for our approach to 3D registration. In particular, Fig.~\ref{fig:alignment} presents the procedure for building and using the 3D alignment graph. See Fig.~\ref{fig:alignment} caption for explanations.

\section{Text grounding in 3D\label{sec:sup_grounding}}
\begin{figure*}[t]
\begin{center}
   \includegraphics[width=1\linewidth]{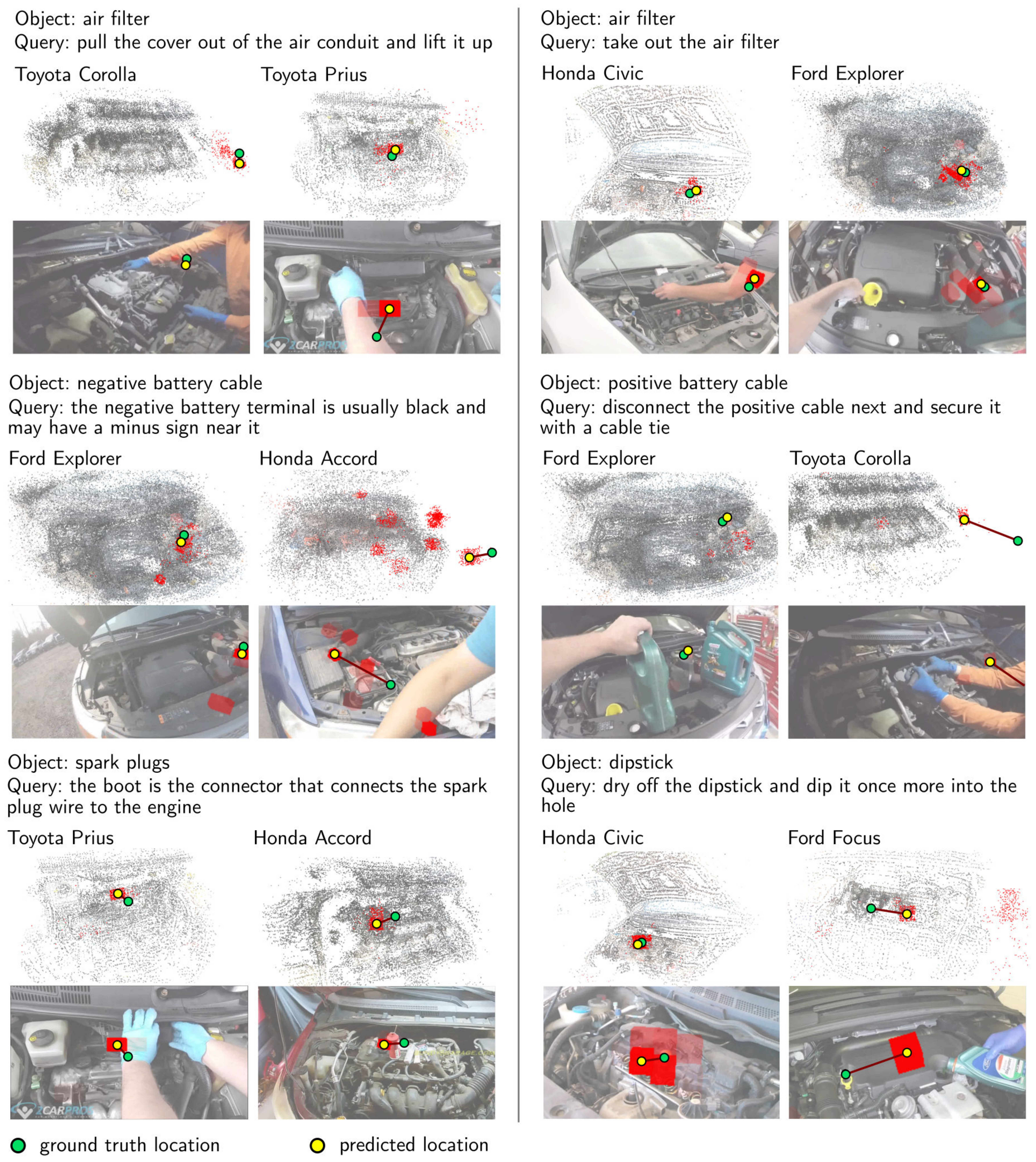}
\caption{{\bf Additional examples of 3D text grounding with our model.} Our model predicts a heatmap over a set of voxels with higher values assigned to locations that correspond to the query text. For clarity, we visualize the predicted voxels in random frames from the videos. The confidence that a given text query corresponds to a given location is shown in \textcolor{red}{red} with higher opacity corresponding to higher confidence. The ground truth 3D locations are shown by \textcolor{gtgreen}{green} dots. Predicted points are shown by \textcolor{yellow}{yellow} dots and correspond to the centers of voxels with the maximum score.
}
\label{fig:text_ground_sup}
\end{center}
\end{figure*}

Fig.~\ref{fig:text_ground_sup} demonstrates additional results of our text grounding similar to the results shown in Sec.~\ref{sec:results} of the main paper. 
Note that our method relies on the underlying 3D model, instead of image appearance. As result, our method is capable to ground queries even when the corresponding object is partially or fully occluded. 

\end{document}